\documentclass[3p,,preprint,12pt]{elsarticle}

\usepackage{lineno}

\usepackage{tabulary,xcolor}
\usepackage{amsfonts,amsmath,amssymb}
\usepackage[T1]{fontenc}
\makeatletter
\let\save@ps@pprintTitle\ps@pprintTitle
\def\ps@pprintTitle{\save@ps@pprintTitle\gdef\@oddfoot{\footnotesize\itshape \null\hfill\today}}
\def\hlinewd#1{%
	\noalign{\ifnum0=`}\fi\hrule \@height #1%
	\futurelet\reserved@a\@xhline}

\AtBeginDocument{\ifNAT@numbers \biboptions{sort&compress}\fi}
\makeatother

\usepackage{ifluatex}
\ifluatex
\usepackage{fontspec}
\defaultfontfeatures{Ligatures=TeX}
\usepackage[]{unicode-math}
\unimathsetup{math-style=TeX}
\else 
\usepackage[utf8]{inputenc}
\fi 
\ifluatex\else\usepackage{stmaryrd}\fi

\usepackage{url,multirow,morefloats,floatflt,cancel,tfrupee}
\makeatletter

\AtBeginDocument{\@ifpackageloaded{textcomp}{}{\usepackage{textcomp}}}
\makeatother
\usepackage{colortbl}
\usepackage{xcolor}
\usepackage{pifont}
\usepackage[nointegrals]{wasysym}
\makeatletter

\def\mcWidth#1{\csname TY@F#1\endcsname+\tabcolsep}

\def\cAlignHack{\rightskip\@flushglue\leftskip\@flushglue\parindent\z@\parfillskip\z@skip}
\def\rAlignHack{\rightskip\z@skip\leftskip\@flushglue \parindent\z@\parfillskip\z@skip}

\usepackage{ifxetex}
\ifxetex\else\if@twocolumn\usepackage{dblfloatfix}\fi\fi

\AtBeginDocument{
	\expandafter\ifx\csname eqalign\endcsname\relax
	\def\eqalign#1{\null\vcenter{\def\\{\cr}\openup\jot\m@th
			\ialign{\strut$\displaystyle{##}$\hfil&$\displaystyle{{}##}$\hfil
				\crcr#1\crcr}}\,}
	\fi
}

\AtBeginDocument{%
	\@ifpackageloaded{endfloat}%
	{\renewcommand\efloat@iwrite[1]{\immediate\expandafter\protected@write\csname efloat@post#1\endcsname{}}}{\newif\ifefloat@tables}%
}%

\def\BreakURLText#1{\@tfor\brk@tempa:=#1\do{\brk@tempa\hskip0pt}}
\let\lt=<
\let\gt=>
\def\processVert{\ifmmode|\else\textbar\fi}

\@ifundefined{subparagraph}{
	\def\subparagraph{\@startsection{paragraph}{5}{2\parindent}{0ex plus 0.1ex minus 0.1ex}%
		{0ex}{\normalfont\small\itshape}}%
}{}

\newcommand\role[1]{\unskip}
\newcommand\aucollab[1]{\unskip}

\@ifundefined{tsGraphicsScaleX}{\gdef\tsGraphicsScaleX{1}}{}
\@ifundefined{tsGraphicsScaleY}{\gdef\tsGraphicsScaleY{.9}}{}
\def\checkGraphicsWidth{\ifdim\Gin@nat@width>\linewidth
	\tsGraphicsScaleX\linewidth\else\Gin@nat@width\fi}

\def\checkGraphicsHeight{\ifdim\Gin@nat@height>.9\textheight
	\tsGraphicsScaleY\textheight\else\Gin@nat@height\fi}

\def\fixFloatSize#1{}
\let\ts@includegraphics\includegraphics

\def\inlinegraphic[#1]#2{{\edef\@tempa{#1}\edef\baseline@shift{\ifx\@tempa\@empty0\else#1\fi}\edef\tempZ{\the\numexpr(\numexpr(\baseline@shift*\f@size/100))}\protect\raisebox{\tempZ pt}{\ts@includegraphics{#2}}}}

\AtBeginDocument{\def\includegraphics{\@ifnextchar[{\ts@includegraphics}{\ts@includegraphics[width=\checkGraphicsWidth,height=\checkGraphicsHeight,keepaspectratio]}}}

\DeclareMathAlphabet{\mathpzc}{OT1}{pzc}{m}{it}

\def\URL#1#2{\@ifundefined{href}{#2}{\href{#1}{#2}}}

\def\UrlOrds{\do\*\do\-\do\~\do\'\do\"\do\-}%
\g@addto@macro{\UrlBreaks}{\UrlOrds}

\edef\fntEncoding{\f@encoding}

\makeatother

\usepackage[english]{babel}
\usepackage[utf8]{inputenc} 
\usepackage[T1]{fontenc}    
\usepackage{hyperref}       
\usepackage{url}            
\usepackage{booktabs}       
\usepackage{amsfonts, mathrsfs, amssymb}       
\usepackage{nicefrac}       
\usepackage{microtype}      
\usepackage[export]{adjustbox}

\usepackage{graphicx}
\usepackage{subfigure,wrapfig}
\usepackage{dsfont}

\usepackage{amsmath}
\usepackage{amsthm}
\usepackage{stmaryrd}
\usepackage{breqn}
\usepackage{shortcuts}

\usepackage{algorithm}
\usepackage{algorithmic}
\usepackage{natbib}
\usepackage[labelfont=bf]{caption}
\usepackage{etoolbox}
\usepackage{subfiles}

\begin{document}



\begin{frontmatter}
	

	\title{Multi-subject MEG/EEG source imaging with sparse multi-task regression}
	
	\author[a3058,a251]{Hicham Janati\corref{c-a987}}
	\ead{hicham.janati@inria.fr}\cortext[c-a987]{Corresponding author.}
	\author[a3058]{Thomas Bazeille}
	\ead{thomas.bazeille@inria.fr}
	\author[a3058]{Bertrand Thirion}
	\ead{bertrand.thirion@inria.fr}
	\author[a200,a251]{Marco Cuturi}
	\ead{cuturi@google.com}
	\author[a3058]{Alexandre Gramfort}
	\ead{alexandre.gramfort@inria.fr}
	\address[a3058]{Université Paris-Saclay, Inria, CEA\unskip, France}
	\address[a200]{Google\unskip, France}
	\address[a251]{ENSAE / CREST\unskip, France}
	
\begin{abstract}
	Magnetoencephalography and electroencephalography (M/EEG) are non-invasive modalities that measure the weak electromagnetic fields generated by neural activity. Estimating the location and magnitude of the current sources that generated these electromagnetic fields is a challenging ill-posed regression problem known as \emph{source imaging}. When considering a group study, a common approach consists in carrying out the regression tasks independently for each subject. An alternative is to jointly localize sources for all subjects taken together, while enforcing some similarity between them. By pooling all measurements in a single multi-task regression, one makes the problem better posed, offering the ability to identify more sources and with greater precision. The Minimum Wasserstein Estimates (MWE) promotes focal activations that do not perfectly overlap for all subjects, thanks to a regularizer based on Optimal Transport (OT) metrics. MWE promotes spatial proximity on the cortical mantel while coping with the varying noise levels across subjects.
	On realistic simulations, MWE decreases the localization error by up to 4 mm per source compared to individual solutions. Experiments on the Cam-CAN dataset show a considerable improvement in spatial specificity in population imaging.
	Our analysis of a multimodal dataset shows how multi-subject source localization closes the gap between MEG and fMRI for brain mapping.
\end{abstract}

	
\end{frontmatter}
\section{Introduction}
\label{s:introduction}
Magnetoencephalography (MEG) measures the magnetic field surrounding the head, while Electroencephalography (EEG) measures the electric potential at the surface of the scalp. Both can do so with a temporal resolution of less than a millisecond~\cite{baillet2017,hari-puce:17}.  Localizing the underlying neural activity at the origin of the signals is a linear inverse problem known as \emph{source imaging}~\cite{Baillet-etal:01,becker-etal:2015,michel-etal:2004,wipf-nagarajan:2009}. From a statistical perspective, it can be regarded  as a \emph{linear regression problem} in high dimension. This linearity follows directly from Maxwell equations. However, this inverse problem is inherently ``ill-posed'': Indeed, the number of potential sources is larger than the number of MEG and EEG sensors, which implies that, even in the absence of noise, different neural activity patterns could result in the same electromagnetic field measurements. This fact makes M/EEG source imaging particularly challenging in the presence of multiple simultaneous active regions in the brain.

To limit the set of possible solutions, prior hypotheses on the nature of the source distributions are necessary. The minimum-norm estimates (MNE) for instance are based on $\ell_2$ Tikhonov regularization which leads to a linear solution~\cite{Hamalainen1994}. An $\ell_1$ norm penalty was also proposed by \citet{mce}, modeling the underlying neural pattern as a sparse collection of focal dipolar sources, hence their name ``Minimum Current Estimates'' (MCE). These methods have inspired a series of contributions in source localization techniques relying on noise normalization such as dSPM~\citep{dspm} and sLORETA~\citep{dspm,sloreta} to correct for the depth bias~\citep{depthbias} or block-sparse norms such as MxNE~\citep{strohmeier-etal:16} and TF-MxNE~\citep{gramfort-etal:2013} to leverage the spatio-temporal dynamics of MEG signals. If other imaging data are available such as fMRI~\cite{liu-etal:06,Ou-etal:10} or diffusion MRI~\cite{deslauriersgauthier-etal:17}, they can be used as prior information for example in hierarchical Bayesian models~\cite{sato-04}.
While such techniques have had some success, source estimation in the presence of complex multi-dipole configurations remains a challenge. To address it, one idea is to leverage the anatomical and functional diversity of multi-subject datasets to improve localization results.

This idea of using multi-subject information to improve the spatial accuracy of M/EEG source imaging has been proposed before in the neuroimaging literature. \citet{larson14} hypothesized that different anatomies across subjects allow for point spread functions that agree on a main activation source but differ elsewhere. Averaging across subjects thereby increases the accuracy of source localization. On fMRI data, \citet{varoquaux11} proposed a probabilistic dictionary learning model to infer activation maps jointly across a cohort of subjects. A similar idea lead \citet{litvak-friston:2008} to propose a Bayesian hierarchical model to cope with inter-subject functional variability. Their model however relied on a common source space for all subjects. This assumption was relaxed by \citet{gala} who proposed a similar Bayesian model which outperformed single subject inverse solvers such as MNE and LORETA on retinotopic data.

Source imaging for a set of subjects can be formulated as solving a set of coupled regression problems. In the statistical machine learning literature such supervised learning problems are commonly referred to as \emph{multi-task} prediction problems. In the M/EEG source imaging literature, to our knowledge, the only contribution formulating the problem as a multi-task regression model employs a Group Lasso with an $\ell_{21}$ block sparse norm~\citep{lim17}. Yet this work forces every potential neural source to be either active for all subjects, or for none of them.


The assumption of perfectly ovelapping functional activity across subjects gets even more unrealistic as we aim for fine spatial resolution in the order of millimeters. In this work, one investigates several multi-task regression models that relax this assumption.
One of them is the multi-task Wasserstein (MTW) model~\citep{janati19a}. MTW is defined through an Unbalanced Optimal Transport (UOT) metric that promotes support proximity across regression coefficients. However, applying MTW to group level data assumes that the signal-to-noise ratio is the same for all subjects. To alleviate this problem, one can infer both the sources and the noise variance for each subject and scale the regularization according to the level of noise. Following similar ideas that lead to the concomitant Lasso~\citep{owen07,zhang12,ndiaye17} or the multi-task Lasso~\citep{massias18a}, the Minimum Wasserstein Estimates (MWE) was first proposed in \citep{janati19b}. However both MTW and MWE rely on convex $\ell_1$ norm penalties which tend to promote sparse solutions at the expense of an amplitude bias. 

One way to mitigate amplitude bias of convex regularizations is to make use of non-convex ones, such as $\ell_p$ pseudo norms with $0 < p < 1$. Due to their non-convexity, these pseudo norms provide better proxies for the $\ell_0$ norm and can thereby promote more sparsity with no amplitude bias~\citep{strohmeier-etal:16,Gorodnitsky_Rao97}. \citet{gasso09} studied a broad family of non-convex penalties and showed that the $\ell_{0.5}$ regularized regression is equivalent to a type of what was previously known as re-weighted or adaptive Lasso \citep{candes2008}.

The paper is organized as follows. In Section~\ref{s:sourceimaging}, one explains how multi-subject M/EEG source imaging can be cast as a multi-task regression problem. Methods from the literature that adopt this approach are briefly recalled. Next some background on UOT metrics are presented. Then the reweighted minimum Wasserstein estimates (MWE$_{0.5}$) method is presented. UOT is proposed to cope with inter-subject spatial variability, $\ell_{0.5}$ pseudo-norm is used to limit amplitude estimation bias and concomitant estimation is exploited to handle subjects affected by different noise levels. Finally experimental results on both simulations and two public MEG datasets~\citep{camcan,ds117} are reported.

A preliminary version of this work was presented at the international conference on Information Processing in Medical Imaging (IPMI)~\citep{janati19b}.

\paragraph{Notation}
We denote by $\mathds 1_p$ the vector of ones in $\bbR^p$ and by $I_n$ the square identity matrix of dimension $n$. $\intset{q}$ denotes the set $\{1, \ldots, q\}$ for any integer $q \in \bbN$. The set of vectors in $\bbR^p$ with non-negative (resp. positive) entries is denoted by $ \bbR^p_+$ (resp. $\bbR^p_{++}$).  On matrices, $\log$, $\exp$ and the division operator are applied elementwise. We use $\odot$ for the elementwise multiplication between matrices or vectors. If $\bX$ is a matrix, $\bX_{i.}$ denotes its $i^{\text{th}}$ row and $\bX_{.j}$ its $j^{\text{th}}$ column. The scalar product between vectors $\bx, \by \in \bbR^p$ is denoted by $\langle\bx, \by\rangle$. We define the Kullback-Leibler (KL) divergence between two positive vectors by $\kl(\bx, \by) = \langle \bx , \log(\bx / \by) \rangle + \langle \by - \bx, \mathds 1_p \rangle$ with the continuous extensions  $0\log(0 / 0) = 0 $ and $0 \log(0) = 0$. We also make the convention $\bx \neq 0 \Rightarrow \kl(\bx | 0) = +\infty$. The entropy of $\bx \in \bbR^n$ is defined as $H(\bx) = - \langle \bx,\log(\bx) - \mathds 1_p \rangle $. The same definition applies for matrices with an element-wise double sum. The $\ell_{21}$ norm of a matrix $A \in \bbR^{p\times S}$ is defined as $\sum_{i=1}^p \|A_{i.}\|_2$. 

 \section{Source imaging as a multi-task regression problem}
\label{s:sourceimaging}

In this section, one shows how the M/EEG inverse problem at the group level can be cast as a multi-task regression problem to estimate jointly source locations and magnitudes for multiple subjects.
\paragraph{Cortically constrained source modeling}
Using a segmentation of the MRI scan of each subject, the positions of potential sources are constructed as a set of coordinates uniformly distributed on the cortical surface~\cite{dspm}. Since synchronized currents flowing along the apical dendrites of cortical pyramidal neurons are thought to be mostly responsible for M/EEG signals~\cite{okada93}, we constrain the dipole orientations to be normal to the cortical surface. Thus, we model the current density as a set of focal current dipoles with fixed positions and orientations. The purpose of source localization is then to infer their amplitudes. The ensemble of possible candidate dipoles forms the \emph{source space}.
\paragraph{Forward modeling}
Let $n$ denote the number of sensors (EEG and/or MEG) and $p$ the number of sources.
Following Maxwell's equations, at each time instant, the electromagnetic fields $\bb \in \bbR^{n}$ are a linear combination of the current density $\bx \in \bbR^p: \textbf{}
\bb= \bL\bx$.
The linear forward operator $\bL \in \bbR^{n\times p}$ is called the \emph{leadfield} or \emph{gain matrix}.
%
However, one observes noisy measurements $\by \in \bbR^n$ given by:
\begin{equation}
\label{eq:inv}
 \by = \bb + \varepsilon  = \bL \bx + \varepsilon \enspace, 
 \end{equation}
where $\varepsilon$ is the noise vector that can be assumed Gaussian distributed $\mathcal{N}(0, \Sigma)$. In practice, $\bL$ is computed by solving Maxwell's equations using for example a Boundary element method~\cite{ha87,mosher-leahy-etal:99,kybic-clerc-etal:05}.
 
\paragraph{Whitening}
Since the signals of MEG sensors are correlated by design, the noise covariance matrix $\Sigma$ is not diagonal. For the inverse problem to be cast as a least squares problem, we perform a whitening transformation of the data. We estimate $\Sigma$ during the baseline (before stimulus) using a cross-validation estimator provided in the MNE software~\cite{engemann15,mne,mne-python}.
Given a noise covariance matrix estimate $\hat{\Sigma}$, the whitening step amounts to computing the transformed data $ \hat{\Sigma}^{-\frac{1}{2}} \by$ and $\hat{\Sigma}^{-\frac{1}{2}} \bL$. In the rest of this paper, we assume that the data are whitened.

\paragraph{Source localization}
Source localization consists in solving in $\bx$ the inverse problem \eqref{eq:inv} which can be cast as a least squares problem:
\begin{equation}
\label{eq:leastsquares}
\bx^\star = \argmin_{\bx \in \bbR^p} \, \|\by - \bL\bx\|^2_2 \enspace .
\end{equation}
%
Since $n \ll p$, problem \eqref{eq:leastsquares} is ill-posed and additional constraints on the solution $\bx^\star$ are necessary. These constraints materialize the type of prior knowledge one has on the source estimates and are commonly applied through $\ell_p$ norms~\cite{haufe-etal:08,ou-etal:2009,strohmeier-etal:16}.
 To favor weak distributed currents over focal sources, one can use a squared $\ell_2$ regularization, leading to minimum-norm Estimates~(MNE)~\cite{Hamalainen1994}: 
  \begin{equation}
 \label{eq:mne}
 \bx^\star = \argmin_{\bx \in \bbR^p} \, \frac{1}{2n} \|\by - \bL\bx\|^2_2 + \lambda\|\bx\|^2_2 \enspace ,
 \end{equation}
where $\lambda > 0$ is a tuning hyperparameter.
Both dSPM \cite{dspm} and sLORETA \cite{sloreta} are variants of MNE that tackle noise normalization.

When analyzing evoked responses, one can instead promote source configurations made of a few focal sources, e.g. using the $\ell_1$ norm. This regularization leads to problem \eqref{eq:lasso} called minimum current estimates (MCE)~\cite{mce}, also known in the machine learning community as the Lasso~\cite{lasso}.
 \begin{equation}
 \label{eq:lasso}
 \bx^\star = \argmin_{\bx \in \bbR^p} \, \frac{1}{2n} \|\by - \bL\bx\|^2_2 + \lambda\|\bx\|_1 \enspace ,
 \end{equation}
 
\paragraph{Depth weighting}
The inverse solutions discussed above are inherently biased towards sources in the superficial layers of the cortex~\cite{lin2006assessing}. Indeed, deep sources require larger amplitude values than superficial ones to produce a magnetic field with similar strength. To circumvent this problem, we normalize the columns of the leadfield $\bL$ by a fraction of their norms. In all our experiments we use a depth weighting of 0.9. Formally, every column $\bL_{.j}$ is normalized by $\|\bL_{.j}\|^{0.9}$~\cite{lin2006assessing,gramfort-etal:2013}.

\paragraph{Common source space}
 Here one proposes to go beyond the classical pipeline and carry out source localization jointly for $S$ subjects. To do so, dipole positions (features) must correspond spatially across subjects. Using an anatomical alignment procedure, the source space of each subject is mapped from an average brain. This process one calls \emph{morphing} uses the sulci and gyri patterns which are matched in an auxiliary spherical inflated cortical surface~\cite{morphing}. The resulting leadfields $\bL^{(1)}, \dots, \bL^{(S)}$ have therefore the same shape $(n\times p)$ with aligned columns; a given column maps to the same brain region across all subjects.
 \paragraph{Multi-task framework}
 Jointly estimating the current density $\bx^{(s)}$ of each subject $s$ can be expressed as a multi-task regression problem where some coupling prior is assumed on $\bx^{(1)}, \dots, \bx^{(S)}$ through a penalty $\Omega$:
 \begin{equation}
 \label{eq:multitask}
    \min_{\bx^{(1)}, \dots, \bx^{(S)} \in \bbR^p} \,
    \frac{1}{2n} \sum_{s=1}^S \|\by^{(s)} - \bL^{(s)}\bx^{(s)}\|^2_2 \,
    + \, \Omega(\bx^{(1)}, \dots, \bx^{(S)}) \enspace.
 \end{equation}
The work of~\citet{lim17} embraces this multi-task framework where the joint penalty is defined through an $\ell_{21}$ mixed norm~\cite{gramfort-etal:2012a}:
\begin{equation}
 \label{eq:multitask-gl}
     \Omega(\bx^{(1)}, \dots, \bx^{(S)}) = \sum_{j=1}^{p} \sqrt{\sum_{s=1}^S {\bx^{(s)}_j}^2} \enspace .
\end{equation}
Following the work of \cite{janati19a}, one can define $\Omega$ using an UOT metric. By doing so, one does not require an exact spatial correspondence between active sources in the group of subjects as enforced by~\cite{litvak-friston:2008,lim17}.

\section{Optimal transport background}
\label{s:ot}

Optimal transport theory offers the mathematical tools to compare probabilities embedded in a metric space~\cite{villani}.
In a discrete space, probability measures can be represented as histograms, i.e a finite dimensional vector of weights. Each entry in such a vector is the mass of a bin in the histogram. In the present context, a bin corresponds to a vertex of the source space, and the weight quantifies the amplitude of the neural activation at this location. In this case, the metric required by optimal transport boils down to the definition of a distance between the bins of the histogram. It is known as the
\emph{ground metric}. For the application considered here, these tools will enable us to define a notion of similarity between source estimates, while taking into account the folded geometry of the cortical mantel.

We now provide some background material on unbalanced OT (UOT), which allows to consider distances between non-normalized histograms. Entries of such histograms do not sum to one, and therefore do not define probabilities. Consider the space $(E, d)$  where each element of $E = \{1, \dots, p\}$ corresponds to a vertex of the cortical source space. Let $\bM$ be the $p\times p$ matrix where the entry $\bM_{ij}$ is the geodesic distance between vertices $i$ and $j$. It encodes the ground metric. \citet{doklady} initially defined an optimal transport distance for normalized histograms (probability measures) on $E$. However, it can easily be extended to non-normalized histograms~\cite{chizat:17,frogner:15}. 
\paragraph{Distances between non-normalized histograms}
Let $\ba, \bb$ be two normalized histograms on $E$. Assuming that transporting a fraction of mass $\bP_{ij}$ from $i$ to $j$ is given by $\bP_{ij} \bM_{ij}$, the total cost of transport is given by $\langle \bP, \bM\rangle = \sum_{ij} \bP_{ij} \bM_{ij}$. To compare $a$ and $b$ one is interested by the minimum amount of mass one needs to move to transport $a$ to $b$. Minimizing this total cost with respect to $\bP$ must be carried out on the set of feasible transport plans with marginals $\ba$ and $ \bb$. It amounts to enforcing that no mass appeared or disappeared during the transport. The (normalized) Wasserstein distance reads:
\begin{equation}
\label{eq:wasserstein}
\text{W}(\ba, \bb) = \min_{\substack{\bP \in \bbR_+^{p\times p} \\ \bP\mathds 1 = \ba, \bP^\top \mathds 1 = \bb}} \, \langle \bP, \bM \rangle \enspace.
\end{equation}

\begin{figure}[!t]
	\begin{minipage}{\linewidth}
	    \centering
		\includegraphics[width=0.8\linewidth]{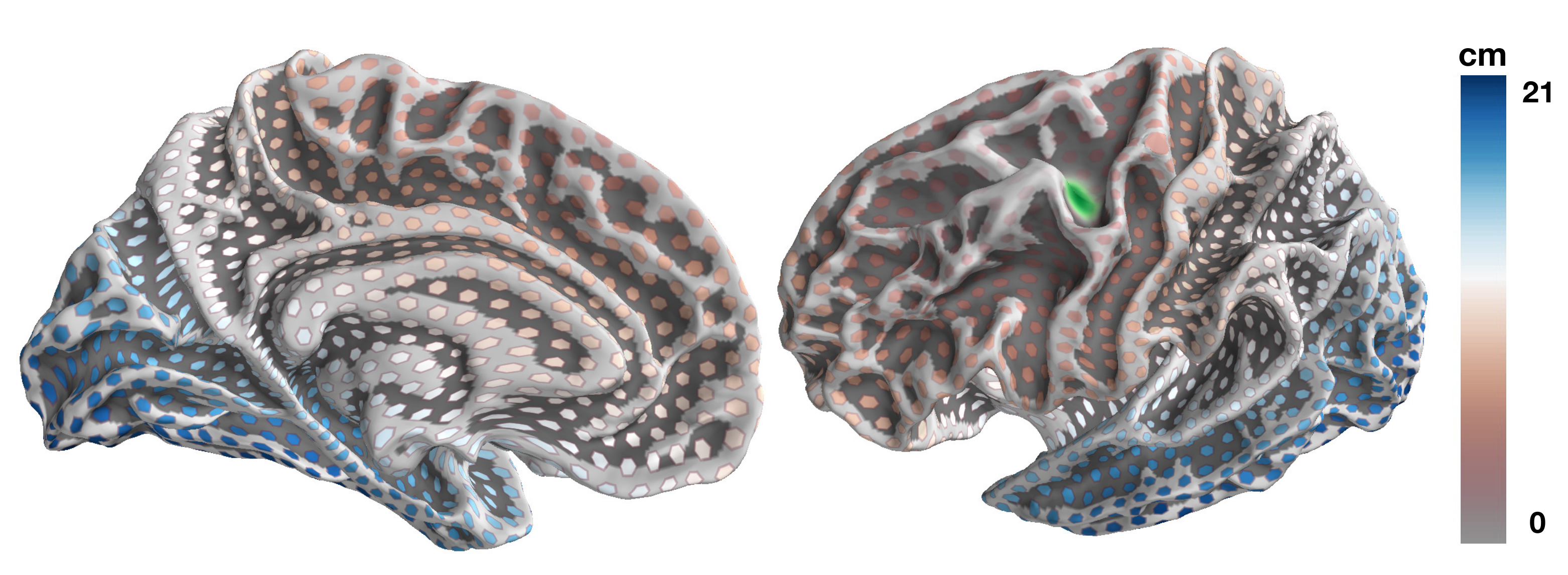}
	\end{minipage}
		\caption{
			Levels of the W distance in cm (a.k.a Earth mover distance) computed between a fixed blurred activation \textbf a (green) and all focal activations \textbf b of the triangular mesh of the cortex.
			\textbf{Left:} Medial view. \textbf{Right:}  Lateral view. \label{f:ot-distance}}
\end{figure}

In practice, if $\ba$ and $ \bb$ are current densities formed each by one focal active source of amplitude one (normalized),  W$(\ba, \bb)$  will quantify the geodesic distance between the two locations along the curved geometry of the cortex. The distance W is however also adapted to multi-dipoles source configurations and possibly spatially extended sources. It is also known as the Earth mover distance (EMD). This property, and the connection with the notion of mass displacement, makes OT metrics adequate for assessing the proximity of functional activations across subjects. We illustrate this behavior of the Wasserstein distance in Figure~\ref{f:ot-distance}. We compute the Wasserstein distances between a fixed brain activation \textbf a spread over 4 vertices (green) and a set of activations \textbf b  concentrated on one vertex. The color of each vertex denotes the Wasserstein distance between \textbf a and \textbf b. If \textbf a was restricted to only 1 vertex, the Wasserstein distance would be equal to the geodesic distance between the pair of activation foci of \textbf a and \textbf b. This distance can be seen as a generalization of the geodesic that compares neural patterns that are spread over multiple vertices of the cortical mesh.

To allow $\ba, \bb$ to be non-normalized, the marginal constraints in \eqref{eq:wasserstein} can be relaxed using a KL divergence:
 \begin{equation}
\label{eq:relaxed-marginals}
\min_{\bP \in {\bbR_+}^{p\times p}} \, \langle \bP, \bM \rangle + \gamma \kl(\bP\mathds 1 | \ba) + \gamma \kl(\bP^\top \mathds 1 | \bb) \enspace,
\end{equation}
where $\gamma > 0$ is a hyperparameter that enforces a fit to the marginals.
\paragraph{Entropy regularization}
Entropy regularization was introduced by \citet{cuturi:13} to propose a faster and more robust alternative to the direct resolution of the linear programming problem \eqref{eq:wasserstein}. Formally, this amounts to minimizing the loss $ \langle \bP, \bM \rangle - \varepsilon H(\bP) $ where $\varepsilon > 0$ is a tuning hyperparameter. This penalized loss function can be written: $\varepsilon \kl(\bP, e^{- \frac{\bM}{\varepsilon}})$ up to a constant~\cite{benamou:15}. Combining entropy regularization with marginal relaxation in \eqref{eq:relaxed-marginals}, we get the unbalanced Wasserstein distance $W_u$ as introduced independently by \citet{chizat:17, frogner:15}:
 \begin{equation}
\label{eq:unbalanced-wasserstein}
W_u(\ba, \bb) = \min_{\bP \in {\bbR_+}^{p\times p}} \,\varepsilon \kl(\bP| e^{- \frac{\bM}{\varepsilon}}) + \gamma \kl(\bP\mathds 1 | \ba) + \gamma \kl(\bP^\top \mathds 1 | \bb) \enspace.
\end{equation}
\paragraph{Generalized Sinkhorn}
Problem \eqref{eq:unbalanced-wasserstein} can be solved as follows.
Let $\bK = e^{- \frac{\bM}{\varepsilon}}$ and $\psi = \gamma / (\gamma + \epsilon)$. Starting from two vectors $\bu, \bv$ set to $\mathds 1$ and iterating the scaling operations $\bu \leftarrow \left(\ba / \bK\bv\right)^\psi$ , $\bv \leftarrow \left(\bb / \bK^\top \bu\right)^\psi$ until convergence, the minimizer of \eqref{eq:unbalanced-wasserstein} can be computed as $ P^{\star} = (\bu_i\bK_{ij}\bv_j)_{i, j \in \intset{p}}$. This algorithm is a generalization of the Sinkhorn algorithm \cite{knopp}. Since it involves matrix-matrix operations, it benefits from parallel hardware, such as GPUs. 
\paragraph{Extension to $\bbR^p$}
Finally, as a M/EEG source estimates can be positive or negative,  we extend the Wasserstein distance $W_u$ to signed measures. We adopt a similar idea to what was suggested in~\cite{mainini,sturm,janati19a} using a decomposition into positive and negative parts, $\ba = \ba^+ - \ba^-$ where $\ba^+ = \max(\ba, 0)$ and $\ba^- = \max(- \ba, 0)$.  For any vectors $\ba, \bb \in \bbR^p$, we define the generalized Wasserstein distance as:
 \begin{equation}
\label{eq:signed-wasserstein}
\widetilde{W}(\ba, \bb) \eqdef W_u(\ba^+, \bb^+) + W_u(\ba^-, \bb^-) \enspace.
\end{equation}
Note that $W_u(\boldsymbol{0}, \boldsymbol{0}) = 0$ (both optimal transport plans must be $\boldsymbol{0}$), thus on positive measures $\widetilde{W} = W_u$.
For the sake of convenience, we refer to $\widetilde{W}$ in \eqref{eq:signed-wasserstein} as the Wasserstein \emph{distance}, even though $\widetilde{W}(\ba, \ba) \neq 0$ in general. In practice, this extension allows to compare source densities across subjects taking into account their polarity. Using common conventions in M/EEG source imaging, positive currents are flowing out of the cortex (from deep cortical layers to superficial ones), while negative currents are flowing into the cortex.
%

\paragraph{Wasserstein barycenters}
As introduced by \citet{agueh:11}, the Wasserstein barycenter $\bxbar$ of a set of inputs $\bx_1,\dots, \bx_S$ is defined as the Fréchet mean of $\widetilde{W}$ across the inputs \cite{agueh:11}:
\begin{equation}
\label{eq:otbar}
\bxbar = \argmin_{\bx  \in \bbR^p}  \frac{1}{S}\sum_{s=1}^{S} \widetilde{W}(\bx^{(s)}, \bx) \enspace,
\end{equation}
In practice, one can write $\bxbar = \bxbar^{(s)_+} - \bxbar^{(s)_-} $. Since the positive and negative parts in \eqref{eq:signed-wasserstein} are not coupled, computing $\bxbar$ only requires a slightly modified version of Generalized Sinkhorn (Algorithm \ref{alg:sinkhorn}) for each part.

\begin{figure}
	\begin{center}
	\includegraphics[width=\linewidth]{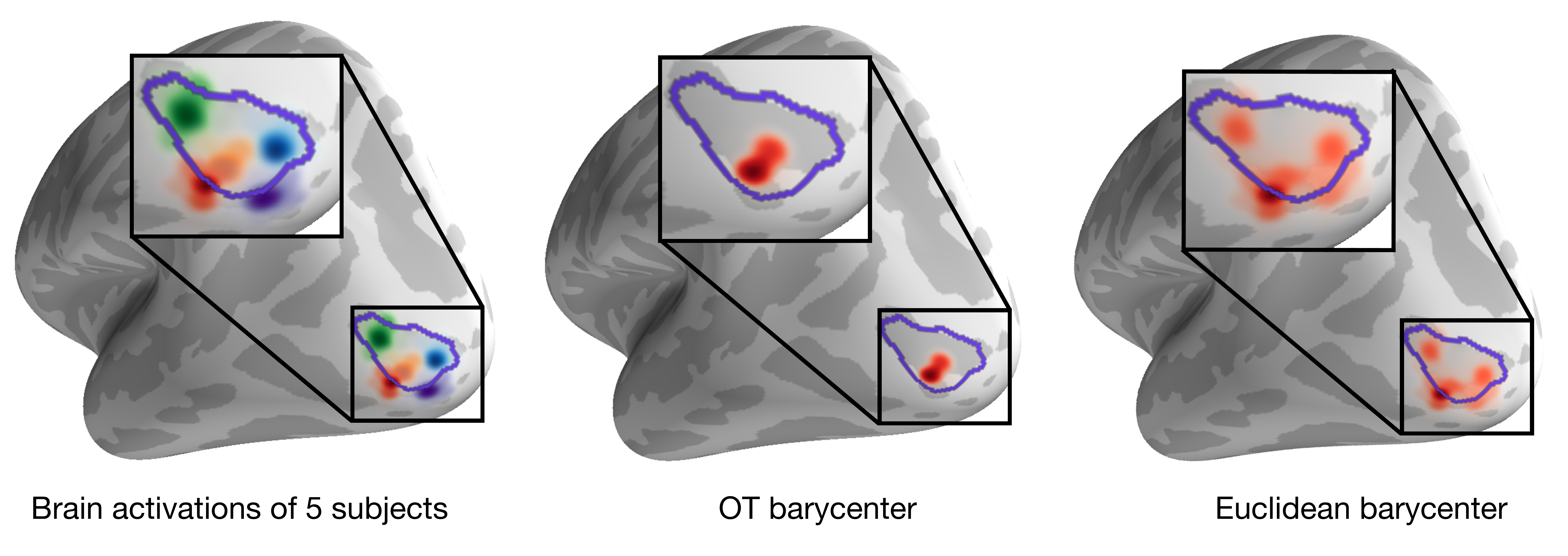}
	\caption{Illustration of the Wasserstein barycenter $\bxbar$ (middle) of 5 activations inputs $\bx^{(s)}$ (left) with random amplitudes between 20 and 30 nAm in the \emph{middle and occipital lunatus sulcus} defined by the \emph{aparc.a2009s} segmentation. $\bxbar$ is located \label{f:ot-bar} at the average location of the inputs with an average amplitude levels. The Euclidean barycenter (right) is the usual mean: it creates undesirable blurring.}
	\end{center}
\end{figure}

Intuitively, one can think of the Wasserstein barycenter as a ``spatial'' averaging technique. Figure \ref{f:ot-bar} illustrates this intuition: the Wasserstein barycenter of the three green activations is located right in their middle location. This idea of averaging source estimates across multiple subjects using Optimal transport  was previously proposed by \citet{gramfort-etal:15}. Their method -- even though based on a different extension of the Earth Mover distance W -- has shown considerable lower smoothing compared to Euclidean averaging and better identification of focal activations. However, it is only carried out as a post-processing step. Here we argue that including optimal transport in the inverse solver upfront allows to improve both the inference of the individual inverse solution of each subject and the average activation across subjects.

\begin{algorithm}
\caption{Generalized Sinkhorn - barycenter computation \citep{chizat:17}}
	\label{alg:sinkhorn}
	\begin{algorithmic}
		\STATE {\bfseries Input:}  $ \bx^{(1)}, \dots, \bx^{(S)} \in \bbR^p_+$
		\STATE {\bfseries Output:} Wasserstein barycenter ($\bxbar$) of $\bx^{(1)}, \dots, \bx^{(S)}$ and marginals $\bm^1, \dots, \bm^{(S)}$.
		\STATE Initialize for $(s = 1, \dots, S) \, (u^{(s)}, v^{(s)}) = (\mathds 1, \mathds 1)$, 
		\REPEAT
		\FOR{$s=1$ {\bfseries to} $S$}
		\STATE $u^{(s)} \gets \left(\bx^{(s)}/Kv^{(s)}\right)^\psi$
		\ENDFOR
		\STATE $\bxbar \gets \left( \tfrac{1}{S}\sum_{s=1}^S (v^{(s)} \odot K^\top u^{(s)})^{1-\psi} \right)^{\frac{1}{1-\psi}} $
		\FOR{$s=1$ {\bfseries to} $S$}
		\STATE $v^{(s)} \gets \left(\bxbar/K^\top u^{(s)}\right)^\psi$
		\ENDFOR
		\UNTIL{convergence}
		\FOR{$t=1$ {\bfseries to} $T$}
		\STATE $\bm^{(s)} = u^{(s)} \odot K v^{(s)}$
		\ENDFOR
	\end{algorithmic}
\end{algorithm}

 \section{Minimum Wasserstein Estimates}
 \label{s:methods}
 \paragraph{The MTW$_q$ model}
 The multi-task Wasserstein model of order $q$ (MTW$_q$), with $0 < q \leq 1$, is the specific case of  \eqref{eq:multitask} with a penalty $\Omega$ promoting both sparsity and spatial proximity between activation foci. The regularization term reads:
 \begin{equation}
    \label{eq:mtw}
    \Omega_{\text{MTW$_q$}}(\bx^{(1)}, \dots, \bx^{(S)}) \eqdef \mu \min_{\bxbar  \in \bbR^p}  \frac{1}{S}\sum_{s=1}^S \widetilde{W}(\bx^{(s)}, \bxbar) + \lambda \|\bx^{(s)}\|_q \enspace,
 \end{equation}
 where $\mu, \lambda \geq 0$ are tuning hyperparameters.
 The minimized OT sum in \eqref{eq:mtw} measures the average distance between all the $\bx^{(s)}$ and their Wasserstein barycenter $\bxbar$. It can thus be seen as quantification of the spatial variability of the source estimates. If $q=1$, one falls back to the MTW model of \citet{janati19a}.
  \paragraph{Minimum Wasserstein Estimates}
  One of the drawbacks of MTW is that $\lambda$ is common to all subjects. Indeed, the loss considered in MTW implicitly assumes that the level of noise is the same across subjects. Following the work of \cite{ndiaye17} on the smoothed concomitant Lasso, we propose to extend MTW by inferring the specific noise standard deviation  ${\sigma^{(s)}}$ along with the regression coefficient $\bx^{(s)}$ of each subject. This allows to scale the weight of the $\ell_q$ regularization according to the level of noise. The Minimum Wasserstein Estimates (MWE$_q$) model reads:
  \begin{equation}
\label{eq:mwe}
\min_{\substack{ \bx^{(1)}, \dots, \bx^{(S)} \in \bbR^p \\ 
	\sigma^{(1)}, \dots, \sigma^{(S)} \in [\sigma_0, +\infty] }} \,  \sum_{s=1}^S \frac{1}{2n  \sigma^{(s)}}\|\by^{(s)} - \bL^{(s)}\bx^{(s)}\|^2_2 \, +\frac{ \sigma^{(s)}}{2} + \, \Omega_{\text{MTW$_q$}}(\bx^{(1)}, \dots, \bx^{(S)}) \enspace,
\end{equation}
where $\sigma_0$ is a pre-defined constant. This lower bound constraint prevents from a null standard deviation and divisions by zero, while also making the feasible set a convex domain. In practice $\sigma_0$ can be set for example using prior knowledge on the variance of the data or as a small fraction of the initial estimate of the standard deviation $\sigma_0 = \alpha \min_s \frac{\| \by^{(s)}\|}{\sqrt{n}}$. In practice we adopt the second option and set $\alpha = 0.01$, although we make sure that it does not affect the solutions by checking that that the estimated $\hat{\sigma}^{(s)}$ are stricly superior to $\sigma_0$.
\begin{algorithm}[b]
	\caption{Reweighted MWE$_{0.5}$}
	\label{alg:reweighting}
	\begin{algorithmic}
		\STATE Initialize weights $\bw^{(s)} = \mathds 1$ for $s=1\dots S$
		\REPEAT 
		\STATE Minimization: solve MWE$_1$ with the weighted $\ell_1$ norms $\|\bw^{(s)}\odot\bx^{(s)}\|_1$
		\STATE Majorization: $\bw_j^{(s)} = \frac{1}{2 \sqrt{|\bx_j^{(s)}|}}$ for all $s, j$
		\UNTIL{convergence}
	\end{algorithmic}
\end{algorithm}

\paragraph{Reweighted Minimum Wasserstein Estimates}
Several studies have shown that non-convex $\ell_q$ with $0 < q < 1$ not only reduce the amplitude bias but also promote a more accurate support estimation~\citep{candes2008, strohmeier15, gasso09}. We define the reweighted Minimum Wasserstein estimates as MWE$_q$ with $q = 0.5$. The resulting optimization problem can be solved in a sequence of weighted instances of MWE$_1$. This reweighting scheme can be seen as a majorization-minimization algorithm~\cite{strohmeier-etal:16}. Indeed, since  the $\ell_{0.5}$ pseudo-norm is separable and concave, it can be upper bounded by its element-wise derivative. The reweighting amounts to iteratively minimizing instances of MWE$_1$ with weighted $\ell_1$ norms and updating the upper bound. These steps are summarized in Algorithm \ref{alg:reweighting}. When some $\bx_j^{(s)} = 0$, the majorization step will cause an overflow error $\bw_j^{(s)} = +\infty$ which corresponds to an infinite amount of regularization and thus leads to $\bx_j^{(s)} = 0$. In practice, one 
can simply filter out the corresponding features or set $\bw_j^{(s)} = \frac{1}{2 \sqrt{|\bx_j^{(s)}| + \eta}}$ where $\eta$ is a small value as proposed by \citet{gasso09}. We adopt this strategy and set $\eta = 10^{-6}$.

 \paragraph{Algorithm for the MWE$_1$ subproblems} 
 We can now explain how to solve the MWE$_1$ subproblems. By combining \eqref{eq:unbalanced-wasserstein}, \eqref{eq:signed-wasserstein}
and \eqref{eq:mwe}, we obtain an objective function taking as arguments $\left((\bx^{(s)+})_s, (\bx^{(s)-})_s, (\bP^{(s)+})_s, (\bP^{(s)-})_s, \bxbar^+, \bxbar^-, (\sigma^{(s)})_s \right)$. This function restricted to all parameters except $(\sigma^{(s)})_s$ is jointly convex \cite{janati19a}. Moreover, each $\sigma^{(s)}$ is only coupled with the variable $\bx^{(s)}$. The restriction on every pair $(\bx^{(s)}, \sigma^{(s)})$ is also jointly convex \cite{ndiaye17}. Thus the problem is jointly convex in all its variables. It can be minimized by alternating optimization. To justify the convergence of such an algorithm, one needs to notice that the non-smooth $\ell_1$ norms in the objective are separable~\cite{Tseng01}.
The update with respect to each $\sigma^{(s)}$ is given by solving the first order optimality condition (Fermat's rule): %
\begin{algorithm}[t]
	\caption{MWE$_1$ algorithm}
	\label{alg:alt}
	\begin{algorithmic}
		\STATE {\bfseries Input:}  $\sigma_0$, $\mu, \epsilon , \gamma, \lambda$ and cost matrix $\bM$. data $(\by^{(s)})_s (\bL^{(s)})_s$.
		\STATE {\bfseries Output:} MWE: $(\bx^{(s)})$, minimizers of \eqref{eq:mwe}.
		
		\REPEAT
		\FOR{$s=1$ {\bfseries to} $S$}
		\STATE Update $\bx^{(s)+}$ with proximal coordinate descent to solve \eqref{eq:cd}.
		\STATE Update $\bx^{(s)-}$ with proximal coordinate descent to solve \eqref{eq:cd}.
		\STATE Update $\sigma^{(s)}$ with \eqref{eq:sigma}.
		\ENDFOR
		\STATE Update  left marginals $\bm^{(1)+}, \dots, \bm^{(S)+}$  and the barycenter $\bxbar^+$ with generalized Sinkhorn of Algorithm \ref{alg:sinkhorn}
		\STATE Update left marginals $\bm^{(1)-} \dots, \bm^{(S)-}$  and the barycenter $ \bxbar^-$ with generalized Sinkhorn  of Algorithm \ref{alg:sinkhorn}
		\UNTIL{convergence}
	\end{algorithmic}
\end{algorithm}
  \begin{equation}
\label{eq:sigma}
\sigma^{(s)} \leftarrow  \frac{ \|\by^{(s)} - \bL^{(s)}\bx^{(s)}\|_2}{\sqrt{n}} \wedge \sigma_0 \enspace,
\end{equation}
 which also corresponds to the empirical estimator of the standard deviation when the constraint is not active. To update the remaining variables, we follow the same optimization procedure laid out in \cite{janati19a} and adapted to MWE in Algorithm~\ref{alg:alt}. Briefly, let $\bm^{(s)+} \eqdef \bP^{(s)+}\mathds 1$ (resp. $\bm^{(s)+} \eqdef \bP^{(s)+}\mathds 1 $), when minimizing with respect to one $\bx^{(s)+}$ (resp. $\bx^{(s)-}$), the resulting problem can be written (dropping the exponents for simplicity):
  \begin{equation}
\label{eq:cd}
\min_{\bx \in \bbR^p_+}  \frac{1}{2n} \|\by - \bL\bx\|_2^2 +  \frac{\mu \gamma}{S} ( \langle \bx, \mathds 1 \rangle - \langle \log(\bx), \bm \rangle ) + \lambda \sigma \|\bx \|_1 \enspace,
\end{equation}
which can be solved using proximal coordinate descent \cite{fercoq}. 
Note that the additional inference of a specific $\sigma^{(s)}$ for each subject allows to scale the Lasso penalty depending on their particular level of noise. 
The final update with respect to $((\bP^{(s)+})_s, (\bP^{(s)-})_s, \bxbar^+, \bxbar^-)$ can be cast as two Wasserstein barycenter problems, carried out using  generalized Sinkhorn iterations \cite{chizat:17}. Note that one does not need to compute the transport plans $P^{(s)}$ since inferring every source estimate $\bx$ only requires the knowledge of the left marginal $\bm = \bP\mathds 1$ which does not require storing $\bP$ in memory.

\section{Experiments}
\subsection{Simulations with semi-real data}
\label{s:results}
\paragraph{Benchmarks}
As discussed in introduction, standard sparse source localization solvers are based on an $\ell_1$ norm regularization, applied to the data of each subject independently. We use the independent Lasso estimator as a baseline. To illustrate the effect of reweighting separately, we also study the performance of a reweighted Lasso, i.e an independent regression with a $\ell_{0.5}$ penalty. Note that reweighted Lasso was coined iterative reweighted mixed-norm (irMxNE) in the context of M/EEG source imaging in \cite{strohmeier-etal:16}. We compare MWE$_{0.5}$ to the Group-Lasso estimator \eqref{eq:multitask-gl} \cite{grouplasso,argyriou-etal:06} which was proposed from multi-subject source imaging to promote functional consistency across subjects~\cite{lim17}. We also evaluate the performance of a more flexible block sparse model where only a fraction of the source estimates are shared across all tasks: Dirty models \cite{dirty}. In Dirty models source estimates are written as a sum of two parts which are penalized with different norms. One is common to all subjects (penalty $\ell_{21}$) and one is specific for each subject (penalty $\ell_1$). We also compare MWE$_{0.5}$ with MWE$_1$ to evaluate the benefits of non-convex penalties. For more details about the compared methods, we refer the reader to the appendix.
%
\paragraph{Hyperparameters of W}
 The parameters defining the Wasserstein distance $\widetilde{W}$ are $\varepsilon$ (entropy regularization) and $\gamma$ (marginal relaxation). Large values of $\varepsilon$ accelerate the convergence of the Sinkhorn algorithm but induce an undesired blurring of the source estimates. Very Low values however lead to numerical instability. We set $\varepsilon$ to $0.002$ divided by the median of the ground metric $M$ which provides a good trade-off between computation speed and sharpness of the barycenter. With the same reasoning, low values of $\gamma$ allow for a ``free'' transport, thus the barycenter converges towards a blurred uniform distribution. We set $\gamma$ to a lower bound $\gamma_{M} = -\frac{\max{M}}{2\log{0.8}} \approx 1$ that guarantees a minimal transport of mass using a strategy proposed in~\cite{janati19a}.

\begin{figure}[!t]
	\centering
    \includegraphics[width=0.6\linewidth]{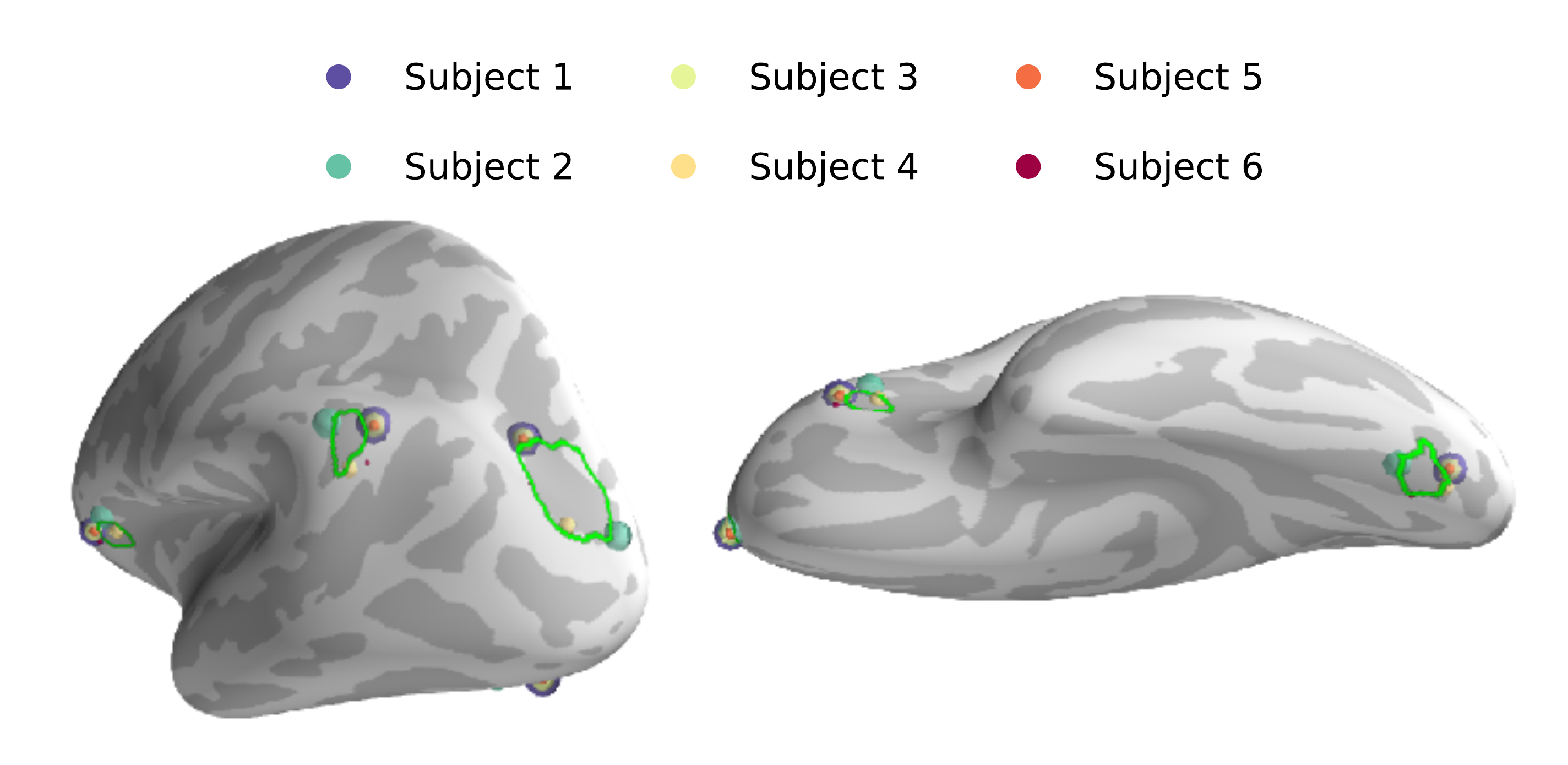}
	\caption{Example of a simulated source configuration with 5 activations for $S=6$ subjects - one activation per label. The 5 labels -- highlighted within green borders -- are taken from the aparc.a2009s FreeSurfer Destrieux parcellation~\cite{destrieux-etal:10}. Different radii are used to distinguish overlapping sources. Here, subjects 1, 3 and 5 share the exact same source locations. \label{f:labels}}
\end{figure}
\paragraph{Simulation data and MEG/fMRI datasets}
In our simulations, we use semi-real data, i.e. we simulate MEG data $\by$ with real leadfield matrices $\bL$. To do so, we rely on the public Cam-CAN dataset \cite{camcan}. We use the MRI scan of each subject to compute a source space and its associated leadfield comprising 2562 sources per hemisphere~\cite{mne}. Keeping only MEG gradiometer channels, we have $n = 204$ observations per subject.
To keep the simulation settings simple, we restrict all leadfields to the left hemisphere. We thus have $S = 32$ leadfields with $p=2562$. We simulate an inverse solution $\bx^{s}$ with 5 sources (5-sparse vector) by randomly selecting one source per label (a.k.a. region of interest) among 5 pre-defined labels using the \emph{aparc.a2009s} parcellation of the Destrieux atlas~\cite{destrieux-etal:10}. To model functional consistency, 50\% of the subjects share sources at the same locations, the remaining 50\% have sources randomly generated in the same labels (see Figure~\ref{f:labels} for an example). Their amplitudes are taken uniformly between 20 and 30\,nAm. Their sign is taken at random with a Bernoulli distribution (0.5) for each label (Hence all subjects share the same polarity of currents in a given label). We simulate $\by$ using the forward model with a covariance matrix $\sigma I_n$. We set $\sigma$ so as to have an average signal-to-noise ratio across subjects equal to 4 (SNR$\eqdef \sum_{s=1}^S \frac{\|\bL^{(s)}\bx^{(s)}\|}{ S\sigma}$).

\begin{figure}
	\includegraphics[width=\linewidth]{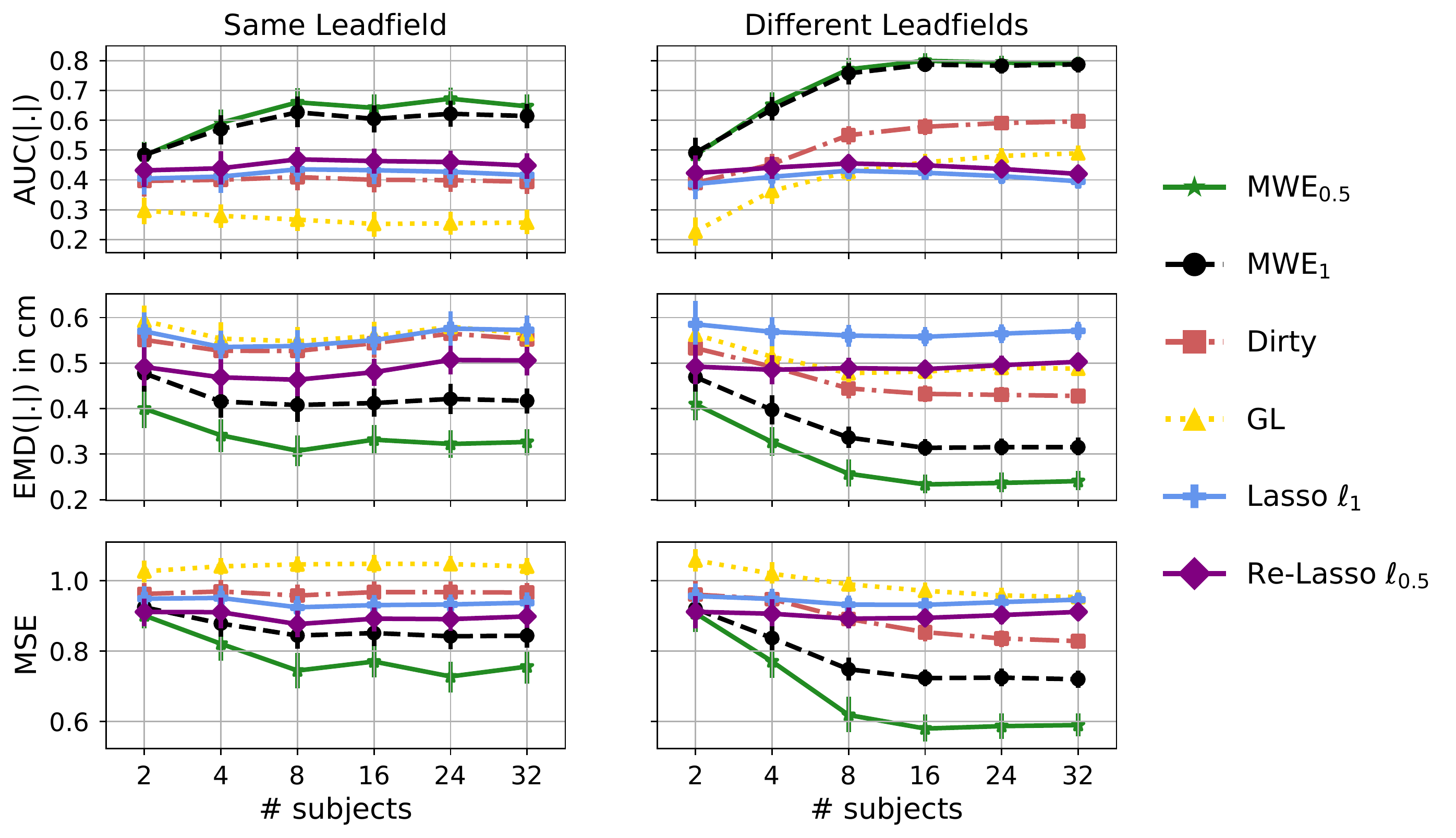}
	\caption{Performance of different models over 30 trials in terms of AUC, EMD and MSE using the same leadfield for all subjects (randomly selected in each trial) (\textbf{left}) and different leadfields (\textbf{right}) computed using Cam-CAN dataset with 5 simulated sources.\label{f:anatomies}}
\end{figure}
We evaluate the performance of all models knowing the ground truth by comparing the best estimates on a grid of hyperparameters in terms of three metrics: the mean squared error (MSE) to quantify accuracy in amplitude estimation, AUC and a generalized Earth mover distance (EMD) to assess supports estimation. We use the PR-AUC (Area under the curve Precision-recall) computed between the absolute values of the coefficients and the true supports. Similarly, the EMD is computed between normalized absolute values of sources. Since $\bM$ is expressed in millimeters, EMD can be seen as an expectation of the geodesic distance between the truth and the source estimates.
For a better intuitive interpretation of the EMD, we compute the EMD per source i.e we divide it by 5. The mean across subjects is reported for all metrics. 
\paragraph{Simulation results}
We vary the number of subjects under two conditions: (1) using the same leadfield for all subjects, as one would do with a template head model, (2) using specific leadfield operators of each subject. Each model is fitted on a grid of hyperparameters and the best AUC/MSE/EMD scores are reported. We perform 30 different trials (with different true activations and noise, different common leadfield for condition (1)) and report the mean within a 95\% confidence interval in Figure \ref{f:anatomies}.

Various observations can be made. First by comparing conditions (1) and (2), one can notice the benefit of using different leadfield operators across subjects. This gain in performance concerns all multi-task models, especially OT based models MWE$_q$. We argue that this improvement is the consequence of the different folding patterns of the cortex across subjects. Indeed, these folding differences lead to different dipole orientations of the same source across subjects, thereby increasing the chances of an accurate localization. Second, note that the Group Lasso~\citep{lim17} performs poorly -- even compared to independent Lasso -- which is expected since simulated sources are not perfectly overlapping for all subjects. Indeed, in the simple case of 2 subjects, on can show that if the fraction of overlapping sources is less than 2/3, Group Lasso performs worse than independent Lasso~\cite{badlinfty}. Our experiments confirm this theoretical result. MWE$_q$ however benefits from the presence of more subjects by leveraging spatial proximity. The mean AUC increases from 0.4 (Lasso) to 0.8. The average error EMD distance is reduced from 6 mm (Lasso) to nearly 2 mm. Finally, even if both MWE$_q$ models show a similar AUC score, the proposed reweighting allows MWE$_{0.5}$ to outperform MWE$_1$ by a significant margin in terms of amplitude estimation (MSE). Finally, by inducing more sparsity, the $\ell_{0.5}$ norm of MWE$_{0.5}$ reduces the number of false positives which are located far from the true sources, thereby reducing the EMD distance by 1mm compared to MWE$_1$.
\subsection{Experiments on MEG data}

\paragraph{Datasets description}
The different strategies were evaluated on two publicly available MEG datasets: DS117~\cite{ds117} and Cam-CAN~\cite{camcan}. DS117 provides MRI, MEG, EEG and fMRI data of 16 healthy subjects to whom were presented images of famous, unfamiliar and scrambled faces. The fusiform face area (FFA) which specializes in facial recognition activates around 170ms after stimulus~\cite{wakeman11, kanwisher1997}. We pick the time point in the contrast response \emph{famous} vs \emph{scrambled} with the peak response for each subject within the interval 150-200ms after stimulus.
Similarly, Cam-CAN provides MEG, EEG and MRI data of around 650 healthy subjects with several types of tasks. We select the youngest 32 subjects (aged between 18 years and 29 years) and use their MEG recordings to study the auditory N100 response. We average the responses of 3 stimuli: 300Hz, 600Hz and 1200Hz with a total of 60 trials. We pick the time point with the peak response within 80-120\,ms after stimulus.
For both datasets, the leadfield operator of each subject was obtained from their T1 MRI scan using a cortically constrained source space formed by about 2500 candidate dipoles per hemisphere.

\paragraph{Model selection}

\begin{figure}[!t]
    \centering
    \includegraphics[width=0.55\linewidth]{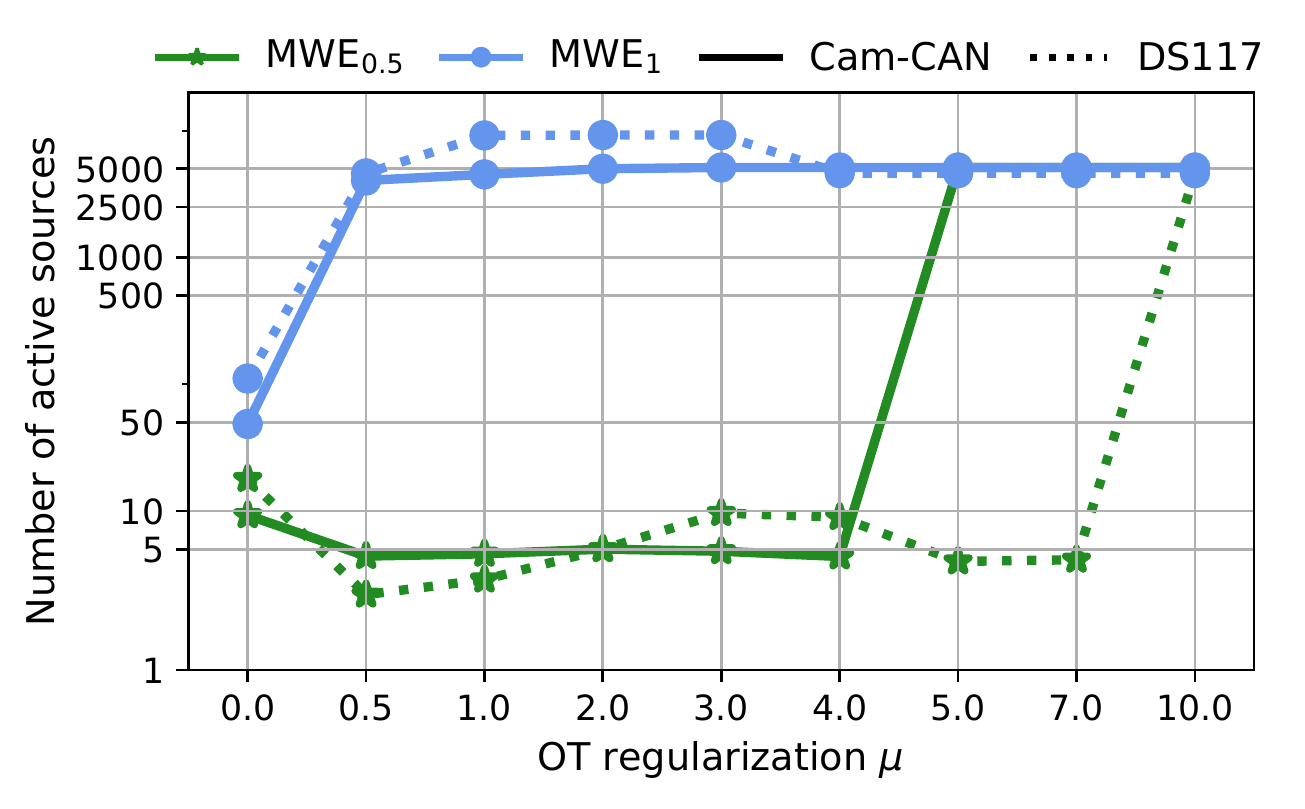}
	\caption{Number of active sources for MWE models with $\lambda = 30\%$. The mean is reported across all subjects. With reweighted MWE, a similar phase transition occurs for both datasets after a certain $\mu_{\max}$. \label{f:model-selection}}
\end{figure}

For all lasso-type models, there exists $\lambda_\max$ such that for $\lambda \geq \lambda_{\max}$ the inverse solution is 0 everywhere. For instance, with $\ell_1$ and $\ell_{0.5}$ we have $\lambda_\max = \frac{\|\bL^\top \by\|_{\infty}}{n}$ \cite{rakotomamonjy19}.
This allows to set $\lambda$ in a relative scale between 0 and 1, making this choice less sensitive to the data. In practice, one can pick a certain value in [0, 1] based on the number of active sources, which is the heuristic used in the following experiments with real data. Even though the choice of $\lambda_\max$ does not theoretically guarantee null source estimates with MWE$_{0.5}$, we observe experimentally that reweighting and the OT regularizer promote even more sparsity with a lower $\lambda$ compared to Lasso models. We use the same relative scaling to set $\lambda$ for MWE$_{0.5}$. The OT regularization parameter $\mu$ controls the level of consistency across subjects. Figure \ref{f:model-selection} shows that for the reweighted MWE$_{0.5}$, there exists a phase transition at a certain value $\mu_\max$, after which the source estimates lose all sparsity and cover the entire cortical mantle uniformly. MWE$_1$ however shrinks the source estimates towards 0 but fails to produce sparse solutions. In practice, based on the complexity of the topographic maps of the MEG data, we select $\lambda$ and $\mu$ that lead to -- on average -- a 2-sparse solution with Cam-CAN ($\lambda = 30\%, \mu=3)$ and a 6-sparse solution with DS117 ($\lambda = 20\%, \mu=0.5$).

\begin{figure}[t]
    \includegraphics[width=\linewidth]{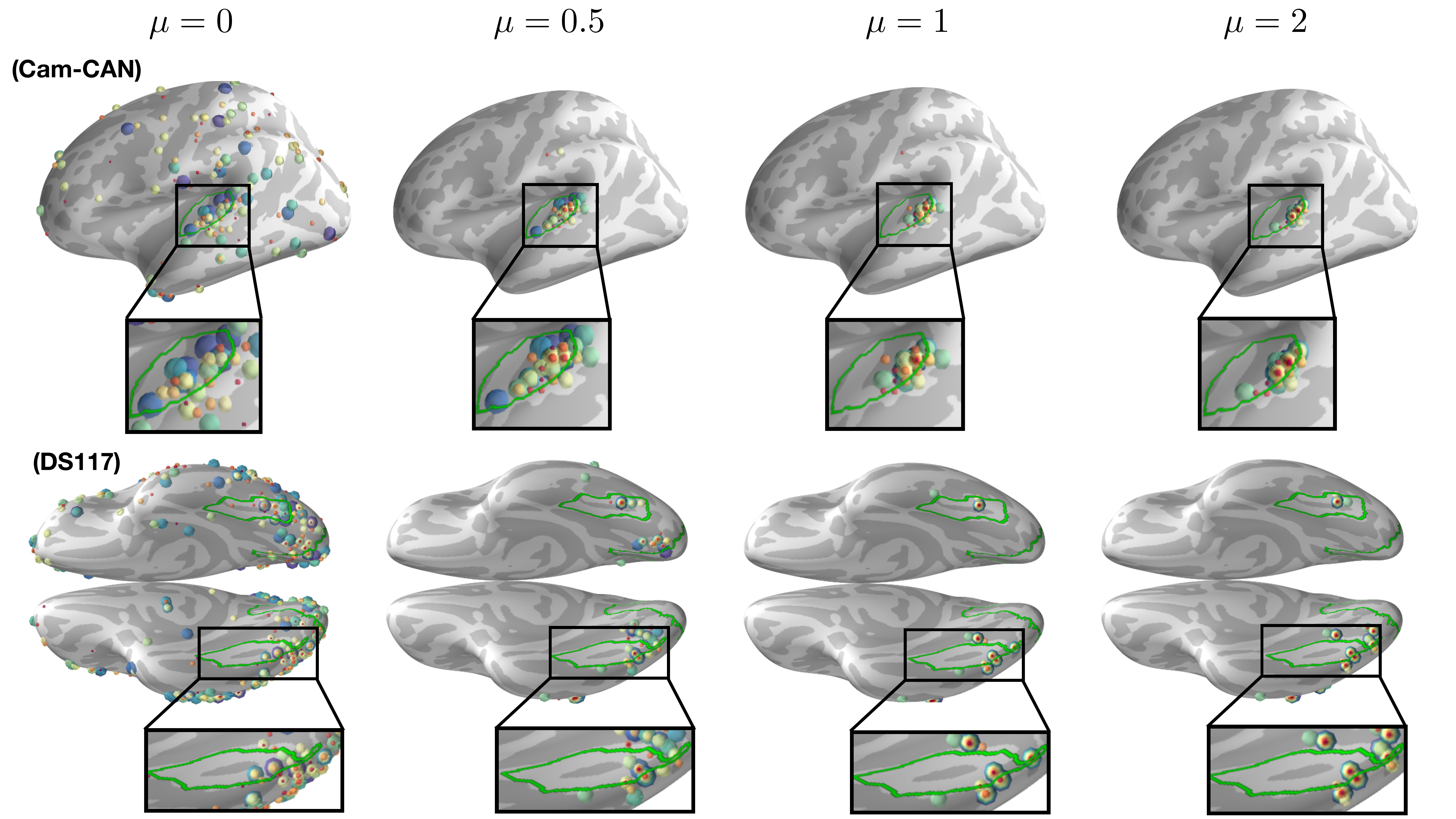}
    \caption{Support of source estimates of MWE$_{0.5}$ recovered in the auditory task of Cam-CAN with 32 subjects (top) and the visual task of DS117 with 16 subjects (bottom). Each color corresponds to a subject. Different radii are displayed for a better distinction of sources. Increasing $\mu$ with $\mu$ < $\mu_{\max}$ promotes functional consistency across subjects. Top: Cam-CAN dataset ($\lambda = 30\%$). Bottom: DS117 dataset ($\lambda = 20\%$).
    \label{f:supports}}
\end{figure}

\begin{figure}[!t]
		\includegraphics[trim={0.cm 0cm 0.cm 0cm}, clip, width=\linewidth]{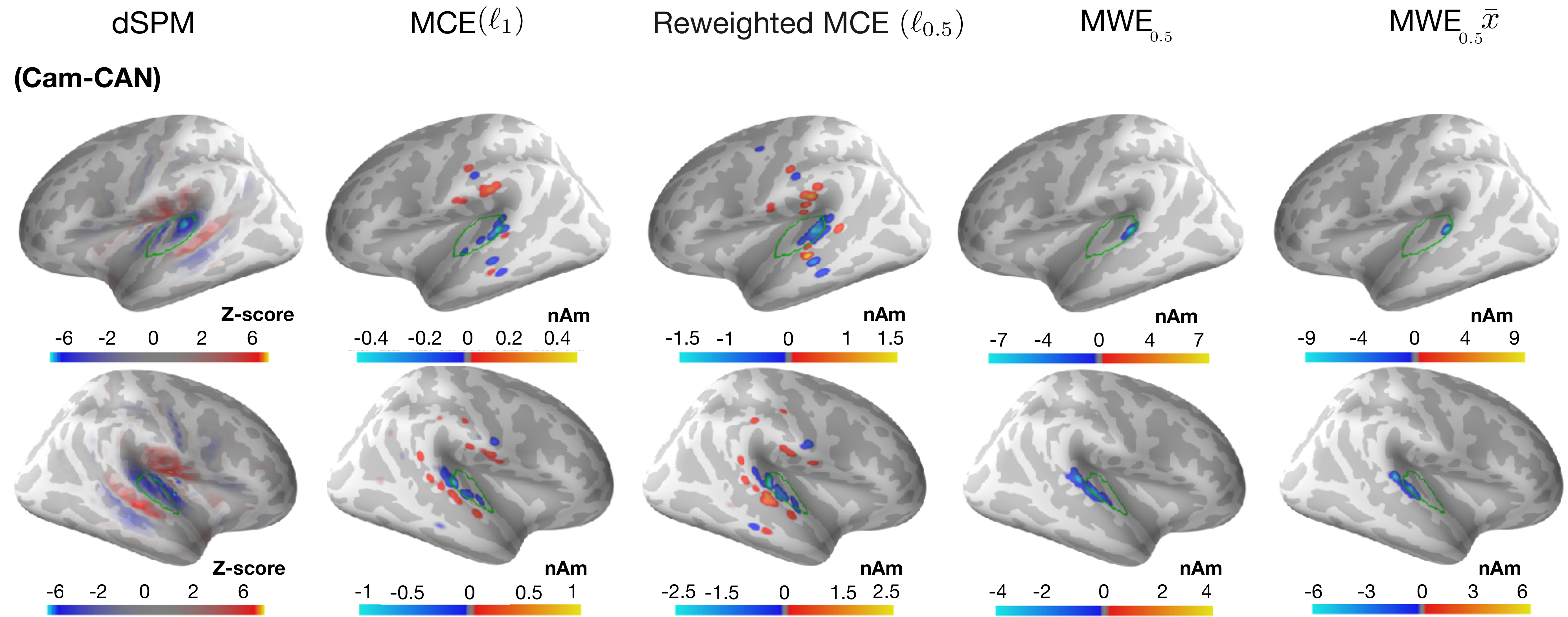}

		\includegraphics[trim={0.cm 0cm 0.cm 0cm}, clip, width=\linewidth]{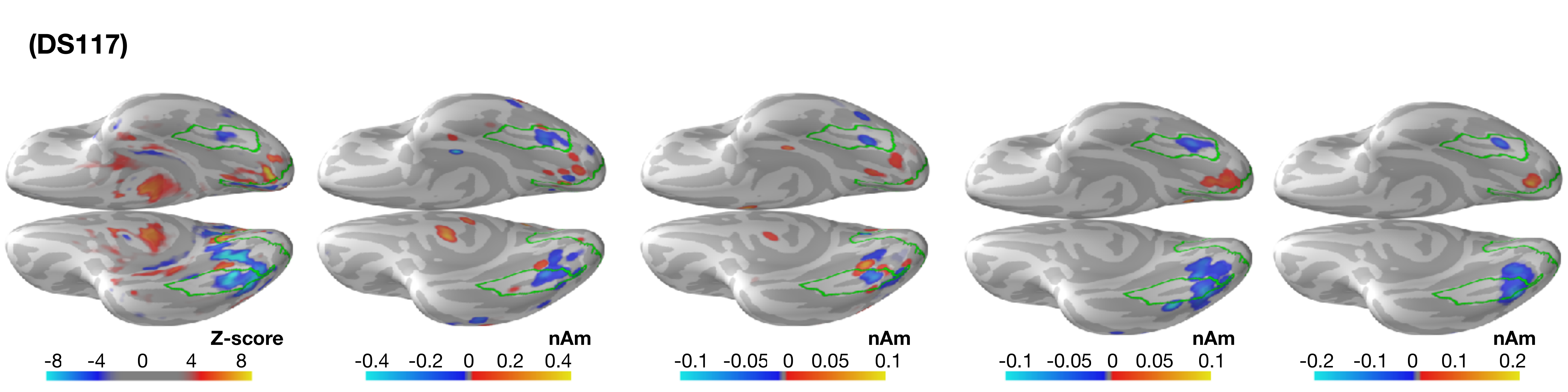}
		\caption{
		Average source estimates of different solvers. \textbf{Top:} Cam-CAN dataset. \textbf{Bottom:} DS117 dataset. MWE$_{0.5}$ $\bxbar$ is the latent variable inferred in the MWE$_{0.5}$ model, corresponding to a Wasserstein barycenter of the MWE$_{0.5}$ source estimates. MWE$_{0.5}$ reduces blurring by promoting functional consistency.  \label{f:averages}}
\end{figure}

\paragraph{MWE for population imaging}
The standard approach to obtain the source estimates from a group of subjects is to average the estimates obtained independently for each subject. Euclidean averaging however induces undesired blurring and sparsity is lost even when the individual solutions are sparse. Figure \ref{f:averages} shows that MWE$_{0.5}$ prevents that from happening. 
Moreover, the latent variable $\bxbar$ of MWE$_{0.5}$ is sharper and more informative at a population level. To compare with single-subject solvers, we compute MCE and reweighted MCE solutions by selecting independently for each subject a $\lambda$ such that the solution is 2-sparse (resp. 6-sparse) for Cam-CAN (resp. DS117). For dSPM, we use the default hyperparameter value 1/SNR$^2$ with SNR\,=\,3. The green borders highlight regions of interest. For Cam-CAN, we use the \emph{neurosynth}~\cite{neurosynth} label corresponding to the \emph{auditory cortex} thresholded at 15 and projected on the surface of the temporal lobe. For DS117, we rely on the \emph{aparc a2009s} segmentation to show both the fusiform gyrus and the primary visual cortex V1. With Cam-CAN, the Euclidean average of the obtained minimum Wasserstein estimates is focal, located right in the auditory cortex. However, The average Lasso and dSPM estimates are dispersed around the auditory cortex with a substantial blurring due to averaging. The visual task of DS117 appears to be the most challenging for several reasons which explain the low amplitude sources. These reasons are discussed in detail in section \ref{s:discussion}.

\begin{figure}[!t]
	\includegraphics[width=\linewidth]{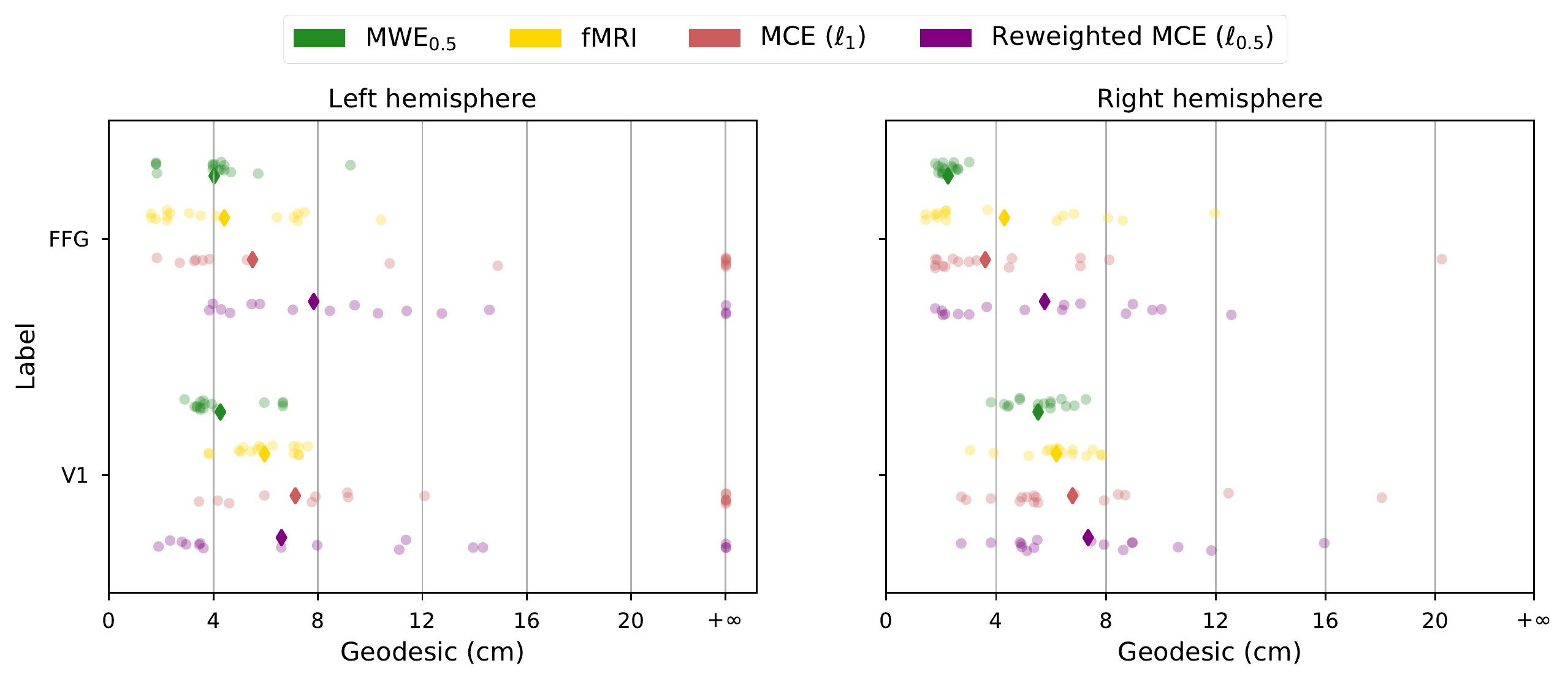}
	\caption{Mean geodesic distance between the mode of the M/EEG derived neural activation map and the vertices of the labels FFA and V1. Each dot represents one of the 16 subjects. For some subjects, MCE / reweighted MCE produce 6-sparse solutions entirely in the right hemisphere, to which the goedesic $+\infty$ is assigned.
		\label{f:geodesic}}
\end{figure}

\paragraph{Comparison with fMRI}
The EEG/MEG inverse problem has an infinite number of solutions. We proposed to regularize it in two ways: (1) at a subject level by favoring focal sources; (2) at a population level by promoting spatial proximity between activation foci. However, one could argue that MWE$_{0.5}$ promotes consistency at the expense of proper fitting of individual data. To address this concern we compute the standardized fMRI Z-score of the conditions \emph{famous vs scrambled faces}. We compare minimum current estimates (MCE or Lasso)~\cite{mce}, reweighted MCE, MWE$_{0.5}$ and fMRI by computing for each subject the mean geodesic distance between the mode of the neural activation map of each subject and the vertices of the Fusiform-gyrus (FFG) as well as the primary visual cortex (V1). Figure~\ref{f:geodesic} shows that the distribution of MWE geodesics is closer to that of fMRI z-maps. By promoting functional similarity, MWE disregards the spurious activation that are far from the regions of interest. Moreover, one can notice that some 6-sparse MCE models cancel out all sources in the left hemisphere (subjects with a geodesic equal to $+\infty$).
\section{Discussion}
\label{s:discussion}


The M/EEG source imaging problem is a notoriously hard inverse problem, in particular when the underlying neural activity is distributed over different coactive brain regions. To tackle this problem, this work proposes to jointly localize sources for a population of subjects by casting the estimation as a multi-task regression problem.

Embracing this formulation of multi-task regression, this work develops three key ideas. First it proposes to use non-linear registration to obtain subject specific leadfield matrices that are spatially aligned. Second it copes with the issue of inter-subject spatial variability of functional activations using Wasserstein metrics and optimal transport theory. Finally, it makes use of advanced techniques from the inverse problem literature using sparsity promoting priors. This allows to model variations of recordings in terms of noise levels using concomitant estimation of sources and noise amplitudes, and it uses $\ell_q$ quasi-norms with $q < 1$ to obtain more accurate source amplitudes.

The classic pipeline of a M-EEG group source imaging study is to perform source localization independently across subjects using inverse solvers such as MNE, MCE, sLORETA, dSPM or MxNE. The group-level analysis is then carried out as a post-processing step by averaging the source estimates of each subject or by aggregating Z-scores in a multiple tests comparison~\cite{takeda19}. This is usually done thanks to a non-linear registration and by averaging of the estimates after mapping them to the same brain template.
In this work, a different approach based on multi-task regression is proposed. The non-linear registration is used to compute leadfield matrices that are spatially aligned. A source space formed by candidate dipoles are defined on the template brain geometry and this source space is warped to individual anatomies for which Maxwell equations are solved numerically. By doing so, we demonstrate improvements in terms of source localization accuracy. This is significant evidence that anatomical variability can be more a blessing than a curse for group level M/EEG source imaging.

This statement is actually inline with the work of \citet{larson14}, who suggested that anatomical differences between subjects can improve the accuracy of the averaged source estimates by emphasizing common sources across subjects. Our simulations confirm this hypothesis not only for averaged estimates but also for individual ones. Indeed, all the multi-task models studied in our simulations improve with more subjects only if the used leadfield operators are different. This striking result suggests that using the same head model in M/EEG group studies for different subjects potentially causes a significant loss of information. One possible explanation of why anatomical differences help is that anatomical variability combined with functional similarities lead to non-redundant information across subjects. Take the example of a shared source across subjects. Different folding patterns of the cortical mantle would lead to different (normal) orientations of the current dipole. Since the relative position of the sensors is not changed, the leadfields -- having different sensitivity maps -- would generate measurements with more information, i.e higher rank. On the contrary, when using the same leadfield for all subjects, an exact same source would lead to similar M/EEG measurements. Quantitatively, our simulations with semi-real data show that multi-subject inverse solvers improve the localization error by almost 4mm per source with different leadfields and only 1 mm when the same leadfield is used. 

By pooling together data from multiple subjects one can increase the number of measurements, hence make the problem less ill-posed.
Yet, this cannot be done without taking into consideration differences between subjects, especially the spatial variability in activation patterns.
To cope with this issue when averaging brain patterns both in M/EEG and fMRI, Wasserstein distances have proven efficient~\cite{gramfort-etal:15}. Through this work,
we explained how they could be included directly in the inverse solver. Thanks to their ability to model spatial proximity between source estimates, the MWE model allows to promote functional similarities across subjects using the geometry of the cortical mantle. Fortunately, the computation of the Wasserstein barycenter does not lead to a computational bottleneck. In our experiments, 40\% to 60\% of time is spent on optimal transport versus proximal coordinate descent. Thanks to careful optimization procedures based on Sinkhorn iterations and block coordinate descent algorithms, the model proposed here runs in a few minutes on empirical M/EEG datasets.

Beyond the use of Wasserstein metrics to cope with spatial misalignments, the proposed MWE$_q$ model brings in two important ingredients from the statistics literature employing sparsity promoting regularizations: concomitant estimation and convex reweighted schemes. By using concomitant estimation, the MWE$_q$ model can cope with the different noise levels and signal-to-noise ratios for the different subjects. This is particularly critical to have the number of hyperparameters of the model that is fixed and does not scale with the number of subjects. In theory, for source imaging with a solver such as dSPM or sLORETA, that is applied independently for all subjects, the regularization parameters could be tuned for each dataset. The MWE$_q$ model has a list of regularization parameters that does not depend on the number of subjects. Besides, results from Figures~4 and 5 demonstrate the benefit of MWE$_{0.5}$ vs. MWE$_{1}$. Employing a more aggressive sparsity promoting regularization improves in particular the source amplitude estimation as shown by the MSE metric. Also it greatly simplifies the setting of the regularization parameter $\mu$ as solutions become suddenly much less sensitive to this choice of parameter.


From a more neuroscientific perspective, the model presented here has potentially interesting consequences. Results on Cam-CAN demonstrate that the barycenters obtained with MWE$_{0.5}$ have a higher spatial specificity. As seen in Figure~7, activation foci in $\bxbar$ are well limited to primary auditory cortices while solvers that are not based on a group-level multi-task regression model lead to spurious activations next to secondary somatosensory cortices and on middle temporal gyrus. On DS117 dataset, the cognitive task performed by the subjects is more advanced, complicating the discussion of the results in terms of localization. Yet, the availability of the fMRI data allows for a quantification of the activation foci between MEG and fMRI. While it is often repeated that fMRI and M/EEG sources are different, and thus brain activation maps obtained by these different modalities should not necessarily match, our findings demonstrate that the proposed method reduces the gap between MEG source imaging and fMRI.


\section*{Appendix}

For the sake of clarity, we provide in this appendix some technical background on all the models discussed in this work.

\subsection*{Adaptive-Lasso}
In its general framework, the adaptive Lasso replaces the $\ell_1$ penalty in \eqref{eq:lasso} with a separable non-convex function. Let $g: \bbR_+ \to \bbR_+$ be a concave differentiable function on $\bbR_{++}$. The adaptive Lasso is defined as:

 \begin{equation}
 \label{eq:adalasso}
 \bx^\star = \argmin_{\bx \in \bbR^p} \, \frac{1}{2n} \|\by - \bL\bx\|^2_2 + \lambda\sum_{j=1}^p g(|\bx_j|) \enspace .
 \end{equation}
 
\citet{gasso09} showed that problem \eqref{eq:adalasso} can be solved iteratively using nested weighted Lasso problems illustrated in algorithm \ref{alg:adalasso-reweighting}. Taking $g:x \to \sqrt{x}$ leads to the $\ell_{0.5}$ reweighting presented in section \ref{s:methods}.

\begin{algorithm}[b]
	\caption{Adaptive Lasso reweighting}
	\label{alg:adalasso-reweighting}
	\begin{algorithmic}
		\STATE Initialize weights $\bw^{(s)} = \mathds 1$ for $s=1\dots S$
		\REPEAT 
		\STATE Minimization: solve Lasso with the weighted $\ell_1$ norms $\|\bw^{(s)}\odot\bx^{(s)}\|_1$
		\STATE Majorization: $\bw_j^{(s)} = g'(\bx_j^{(s)})$ for all $s, j$
		\UNTIL{convergence}
	\end{algorithmic}
\end{algorithm}

\subsection*{Group Lasso}
The Group Lasso \cite{grouplasso, lim17} solves several related regression jointly by enforcing a form of structured sparsity. In the context of source imaging, this structured sparsity corresponds to a strict consensus across subjects to decide whether a source is active or not. Formally, the group lasso solves:

 \begin{equation}
 \label{eq:grouplasso}
    \min_{\bx^{(1)}, \dots, \bx^{(S)} \in \bbR^p} \,
    \frac{1}{2n} \sum_{s=1}^S \|\by^{(s)} - \bL^{(s)}\bx^{(s)}\|^2_2 \,
    + \, \sum_{j=1}^{p} \sqrt{\sum_{s=1}^S {\bx^{(s)}_j}^2}  \enspace.
 \end{equation}
The double sum penalty can be seen as an $\ell_1$ penalty applied to a vector of $\ell_2$ norms taken across subjects: only some $\ell_2$ norms are non-zero. Therefore, source are cancelled out for all subjects or for none of them.

\subsection*{Dirty models}
The assumption of identical sources for all subjects is clearly not realistic, Dirty models~\citep{dirty} relax this assumption by decomposing the source vector of each subject $s$ into two parts: $\bx^{(s)} = \bx_c^{(s)} +  \bx_s^{(s)}$, where the support of $\bx_c^{(s)}$ is common to all subjects and $\bx_s^{(s)}$ is specific to each one. The regularization
then writes:

\begin{equation*}
	\label{eq:dirty-penalty}
    J_{\text{Dirty}}(\bx_c, \bx_s) = \mu \|(\bx_c^{(1)}, \dots, \bx_c^{(s)})\|_{21} + \lambda \sum_{s=1}^{S} \|\bx_s^{(s)}\|_1 \enspace .
\end{equation*}
Dirty models hence aim at solving the optimization problem:

\begin{equation*}
	\label{eq:dirty}
    \min_{\bx_c, \bx_s} \frac{1}{2n} \sum_{s=1}^S\|\bL^{(s)}(\bx_c^{(s)} + \bx_s^{(s)}) - \by^{(s)}\|_2^2 + J_{\text{Dirty}}(\bx_c, \bx_s) \enspace .
\end{equation*}

When $\bx_s = \boldsymbol{0}$ (resp. $\bx_c = \boldsymbol{0}$) one falls back to a Group Lasso (resp. independent Lasso). Indeed, the $\ell_{21}$ norm forces the $\bx_c$ to share the same active locations across subjects. The advantage of Dirty models over the Group Lasso is that it is agnostic with respect to the degree of similarities across subjects. 



\section*{Acknowledgements}
\noindent
This work was funded by the ERC Starting Grant
SLAB ERC-YStG-676943 and a \emph{chaire d’excellence de l’IDEX Paris Saclay}.

\section*{References}
\bibliographystyle{apa}
\bibliography{references}

\end{document}